# HIGH RESOLUTION SURFACE RECONSTRUCTION OF CULTURAL HERITAGE OBJECTS USING SHAPE FROM POLARIZATION METHOD

F. S. Mortazavi [1]*, M. SaadatSeresht [2]

[1] Institute of Cartography and Geoinformatics, Leibniz University Hannover, Hannover, Germany – (Faezeh.Mortazavi@ikg.uni-hannover.de)
[2] School of Surveying and Geospatial Engineering, College of Engineering, University of Tehran, Tehran, Iran – (msaadat@ut.ac.ir)

**Commission II**

**ABSTRACT:**

Nowadays, three-dimensional reconstruction is used in various fields like computer vision, computer graphics, mixed reality and digital twin. The three- dimensional reconstruction of cultural heritage objects is one of the most important applications in this area which is usually accomplished by close range photogrammetry. The problem here is that the images are often noisy, and the dense image matching method has significant limitations to reconstruct the geometric details of cultural heritage objects in practice. Therefore, displaying high-level details in three-dimensional models, especially for cultural heritage objects, is a severe challenge in this field. In this paper, the shape from polarization method has been investigated, a passive method with no drawbacks of active methods. In this method, the resolution of the depth maps can be dramatically increased using the information obtained from the polarization light by rotating a linear polarizing filter in front of a digital camera. Through these polarized images, the surface details of the object can be reconstructed locally with high accuracy. The fusion of polarization and photogrammetric methods is an appropriate solution for achieving high resolution three-dimensional reconstruction. The surface reconstruction assessments have been performed visually and quantitatively. The evaluations showed that the proposed method could significantly reconstruct the surfaces' details in the three-dimensional model compared to the photogrammetric method with 10 times higher depth resolution.

## 1. INTRODUCTION

Today, three-dimensional reconstruction is used in various fields such as computer vision, computer graphics, mixed reality and digital twin and many applications such as cultural heritage, medicine and industry. The definition of three-dimensional reconstruction in computer graphics and computer vision is the process of obtaining the shape and appearance of objects. This process can be done by active or passive methods (Moons et al., 2010).

The resolution of three-dimensional models of photogrammetric cameras may not be sufficient for high-resolution applications such as the three-dimensional reconstruction of cultural heritage objects. The problem here is that the images are often noisy, and the dense image matching method has significant limitations to reconstruct the geometric details. One of the best solutions for high-resolution three-dimensional reconstruction is to combine a three-dimensional model with the details obtained from active methods such as photometric stereo (PS) (Esteban et al., 2008; Park et al., 2013; Zhou et al., 2013; Haque et al., 2014), shape from structured light (Nehab et al., 2005), or shape from shading (Wu et al., 2011; Oxholm and Nishino, 2014; Langguth et al., 2016). The initial three-dimensional model maintains the global and geometric structure of the object in this combined solution. On the other hand, the active methods provide the details needed for high resolution three-dimensional reconstruction.

Active three-dimensional reconstruction methods have some drawbacks, for example, their need for a light source and proximity to the object. If this combination can be done passively, we will see significant progress in improving the accuracy of the photogrammetric methods in the future. Hence, we consider doing this by using a passive method.

In this paper, to achieve a high-resolution surface reconstruction, we implement a passive combined three-dimensional reconstruction method that does not have the drawbacks of the active methods. In this approach, the output of a polarization-based method and a conventional photogrammetric method such as multi-view stereo (MVS) are combined to achieve high resolution surface reconstruction.

The concept of polarization method is based on changes in the polarization of reflected light from the objects. These changes can be analyzed by locating a polarizing filter in front of the digital camera and rotating the filter (Wolff, 1997; Atkinson and Ernst, 2018) or using a polarization camera (Polarization_camera, 2020; Yang et al., 2018). These images can obtain the polarization information for each pixel, and the surface normal can be reconstructed with high resolution, which results in local surface reconstruction with details.

The polarization method is based on Fresnel equations (Collett, 2005) and has potential advantages over conventional photogrammetric methods.

- Passive method: Unlike active methods such as structured light, photometric stereo or shape from shading, which usually use active light, the Polarization method can work without the requirement of active or calibrated lighting.
- different lighting conditions: Object reconstruction using SfP (Shape from Polarization) can be used in different lighting conditions such as indoors, outdoors and even with natural light.

---
\* Corresponding author

- different materials: This method can reconstruct objects of different materials, both dielectric and non-dielectric, shiny and glassy objects.

The SfP method is not yet complete and mature, and it still has some limitations and challenges (Mitra and Nguyen, 2003; Atkinson and Hancock, 2006).

- Phase ambiguity: Using the polarization method alone causes ambiguity in the phase angle. This ambiguity can be a significant problem since the azimuth angle reconstructs the normal vectors. This ambiguity occurs because the linear polarizing filter cannot distinguish azimuth angles, shifted apart by π radians.
- Noise: The polarization method is susceptible to fronto-parallel geometries, when the zenith angle is close to zero. Since the polarizing filter reduces the intensity of the light, the image capturing should not be in a way that amplifies the noise of the method.
- Refractive index: In this method, for three-dimensional shape reconstruction, it is necessary to know the refractive index of objects.

In this paper, we present a combined solution to increase depth resolution using shape from polarization method. Experimental results show that our method can be used as an aided technique in photogrammetry to overcome the weaknesses of conventional photogrammetric methods in reconstruction of surface details.

## 2. PRIOR WORKS

In recent years, many kinds of research have been conducted on three-dimensional reconstruction based on the polarization of light. In this section, some of the essential researches are introduced.

There are several methods in which reconstruction is done using only the polarization method. In the SfP method, the information (degree of polarization, phase angle, zenith angle, Etc.) can be obtained from the polarized images to recover the surface normal. This method can be used in various polarized models such as dielectric or non-dielectric, diffuse or specular and hybrid models. Reconstructed surfaces using polarization information have ambiguities, so methods such as convexity of objects and surface normals along the boundary can resolve the ambiguity (Miyazaki et al., 2003; Atkinson and Hancock, 2006). The shape from polarization method has applications to specular surfaces, for example, surface reconstruction of specular metallic objects (Morel et al., 2005). Huynh et al. (2010) described an approach for dielectric surfaces that solved the problem of simultaneous shape and refractive index recovery while assuming convexity to disambiguation. Miyazaki et al. (2002) presented a method for obtaining the surface orientation in transparent objects in which they analyzed the degree of polarization at both far-infrared and visible wavelengths.

Considering that using the SfP method alone has some limitations, it can be combined with conventional photogrammetric methods or point clouds obtained from a laser scanner, Kinect, Etc. Since the photometric constraints complement polarization constraints, the resolution of the depth maps can be dramatically increased. For example, in black specular objects, the surface normal and three-dimensional coordinates can be estimated by multi-view space carving (Miyazaki et al., 2012 and 2016). Drbohlav and Sara (2001) used photometric stereo to reduce ambiguity with uncalibrated lights to design two constraints that shows normals projections on planes perpendicular to the direction of viewing and illumination. Furthermore, to resolve ambiguity (Ngo Thanh et al., 2015), two constraints on shading and polarization have been used to estimate surface normal, refractive index, and light directions under different lighting. The method presented by Atkinson and Hancock (2007b) combines polarization, shading, and stereo information which causes an object to be divided into patches and simplify matching. Mahmoud et al. (2012) and Smith et al. (2016) presented a direct method for shape reconstruction without nonlinear optimization, which uses both the polarization and the shading methods and assumes constant orthographic projection and albedo. Following up on the previous work, Tozza et al. (2017) presented a differential approach using SfP and shape from shading to estimate surface depth directly with several constraints. In another work, Mecca et al. (2017) proposed a method that derives an albedo independent and a differential model based on a system of hyperbolic PDEs. Smith et al. (2018) developed their last work (Smith et al., 2016), which shows that either illumination or albedo can be estimated from polarization data. A similar method has been described by Yu et al. (2017), which uses a nonlinear least-squares optimisation that minimises the residuals in all pixels between observed and predicted intensities, in contrast to previous methods. A novel approach has been presented by Atkinson (2017) to combine the polarization information with photometric stereo data and resolve ambiguities using a region growing process. Another combined method using SfP and PS methods has been proposed by Atkinson and Hancock (2007a), where polarization data in different light sources estimate the surface normal of the dielectric surfaces, and then the PS method is used for disambiguation. The integration of multi-view stereo with polarization information has been used for feature correspondence matching (Atkinson and Hancock, 2005). In the following, an efficient method for three-dimensional reconstruction of featureless areas has been proposed by Cui et al. (2017) that combines the multi-view stereo with the photometric information from polarization. Kadambi et al. (2017) used the normal surface of the polarization method to improve the depth maps so that the surface normal is combined with depth maps. Zhu and Smith (2019) proposed a technique that combines polarization, stereo, and shading uses a depth map as a guide surface to solve phase ambiguity.

The previous conventional methods for SfP have been completely physics-based. Ba et al. (2020) have made the first attempt to use deep learning and CNN networks to obtain normals from polarization. This proposed approach has achieved more reliable results compared to previous conventional methods. Kalra et al. (2020) proposed a solution for segmenting transparent objects using the SfP method and deep learning. Deschaintre et al. (2021) presented a novel method that uses deep learning to solve SfP problems and achieve depth and surface normal in which it uses frontal flash illumination. In the latest research on this topic, (Lei et al., 2021) has presented a learning-based framework that uses a real-world dataset for reconstruction.

This paper presents a combined three-dimensional reconstruction method based on polarization information to enhance depth maps' resolution.

## 3. PROPOSED METHOD

The implementation of the proposed method is made up of two main steps, as follow:
1. Obtaining high resolution height from polarization method.
2. Combining the height obtained from polarization method with an initial high accuracy depth map from photogrammetry.

In the following, the details of each step will be discussed.

### 3.1 Height from polarization

The polarization method is a physically-based method that uses Fresnel equations to obtain polarization information. Also, in this method, the type of light reflection in each pixel is required. When a surface reflects unpolarized light, it becomes partly polarized (Wolff and Boult, 1993). If this light is reflected directly from the object's surface, it is called specular reflection, and if light penetrates the object and then is reflected due to subsurface scattering, it is called diffuse reflection.

In this method, there are important parameters for surface reconstruction that must be obtained using information from polarization images. One of these parameters is the zenith angle. The angle formed between the surface normal and the light reflected vector from the object is called the zenith angle.

In this paper, we obtain the zenith angle using the degree of polarization of the reflected light. We can capture polarized images of an object using different rotations of a linear polarizing filter which is located in front of the digital camera. The changes in polarized light intensity recorded in the images are sinusoidal relative to the polarizer angles. (See Fig. 1).

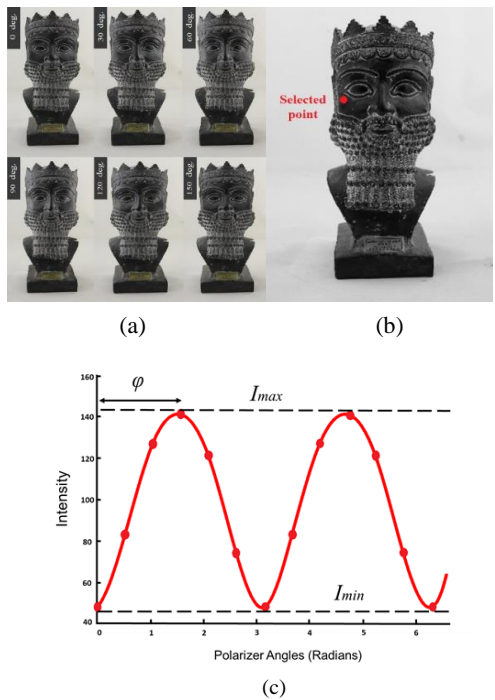

**Figure 1**. (a) Images are captured in different rotations using the linear polarizing filter and a Canon-EOS7D camera; (b) The selected point on the Statue of Dariush and (c) plotting the intensity values for the selected point in each image relative to the polarizer angles, showing that the changes are sinusoidal.

As shown in Fig. 1a, this method requires an image sequence (at least three images). Images are captured at different rotations of the polarizer angle ($\varphi_{POL}$), and the intensity of each pixel varies sinusoidally between $I_{min}$ and $I_{max}$. Hence, the intensity can be written as:

$$I(\varphi_{POL}) = \frac{I_{max} + I_{min}}{2} + \frac{I_{max} - I_{min}}{2} cos(2(\varphi_{POL} - \varphi)) \quad (1)$$

$I_{min}$, $I_{max}$ and $\varphi$, are three unknown variables in this equation that are the lowest and highest values of intensity and phase angle, respectively.

The sinusoidal decomposition leads to obtaining the polarization information at each pixel, and three crucial quantities can be calculated by using them (Wolff, 1997), which are phase angle, $\varphi$, the unpolarized intensity, $i_{un}$, and the degree of polarization, $\rho$, where:

$$i_{un} = \frac{I_{max} + I_{min}}{2} \quad (2)$$

and the degree of polarization can be written as:

$$\rho = \frac{I_{max} - I_{min}}{I_{max} + I_{min}} \quad (3)$$

**3.1.1 Obtaining the azimuth angle:** The azimuth angle of the surface normal $\phi \in [0, 2\pi]$, is the phase angle $\varphi$ after disambiguation. The phase angles obtained from Eq. (1) are ambiguous, and the ambiguity depends on the type of pixel reflection (diffuse or specular). The azimuth angle is given by Eq. (4) for diffuse pixels as well as Eq. (5) for specular pixels:

$$\phi = \varphi \ or \ \varphi + \pi \quad (4)$$

$$\phi = \varphi \pm \frac{\pi}{2} \quad (5)$$

**3.1.2 Obtaining the zenith angle:** The degree of polarization presented in Eq. (3) is based on the amplitude of Eq. (1). It can also be rewritten by substituting the Fresnel equations (Hecht, 2002) in Eq. (3).

$$\rho = \frac{(\eta - \frac{1}{\eta})^2 \ sin^2\theta}{2 + 2\eta^2 - (\eta + \frac{1}{\eta})^2 \ sin^2\theta + 4 \ cos\theta \ \sqrt{\eta^2 - sin^2\theta}} \quad (6)$$

where $\eta$ and $\theta$ are the refractive index and zenith angle, respectively. For diffuse pixels the range of $\eta$ is between 1.4 and 1.6 and in this paper, the value of $\eta$ is assumed to be known.

The zenith angle $\theta \in \left[0, \frac{\pi}{2}\right]$ can be estimated by closed-form, which depends on refractive index and degree of polarization (Smith et al., 2016):

$$f(\rho, \eta) = \sqrt{\frac{2\rho + 2\eta^2\rho - 2\eta^2 + \eta^4 + \rho^2 + 4\eta^2\rho^2 - \eta^4\rho^2 - 4\eta^3\rho\sqrt{-(\rho-1)(\rho+1)} + 1}{\eta^4\rho^2 + 2\eta^4\rho + \eta^4 + 6\eta^2\rho^2 + 4\eta^2\rho - 2\eta^2 + \rho^2 + 2\rho + 1}} \quad (7)$$

Since the specular pixels do not reflect any diffuse light, we cannot use Eq. (6) to obtain the zenith angle of these pixels. So, the degree of polarization for specular pixels is given by:

$$\rho^{spec} = \frac{2 \ sin^2\theta \ cos\theta \ \sqrt{\eta^2 - sin^2\theta}}{\eta^2 - sin^2\theta - \eta^2 \ sin^2\theta + 2 \ sin^4\theta} \quad (8)$$

Depending on the reflection type of each pixel (diffuse or specular), Eq. (6) or Eq. (8) is used to obtain the zenith angle.

The depth estimation problem can be solved using a large linear equation system with the polarization information in each pixel. As mentioned before, the phase angle obtained from Eq. (1) is ambiguous. Smith et al. (2016) expressed this ambiguity as two collinearity conditions in which the phase ambiguity is resolved optimally when the linear system is solved for depth. The first constraint is written in Eq. (9), where a vector in the image plane, [$sin\phi \cos\phi$], is needed. This constraint is satisfied by either of the two possible azimuth angles:

$$n(u) \cdot [cos(\phi(u)) \ -sin(\phi(u)) \ 0]^T = 0 \quad (9)$$

in this equation, $u = (x, y)$ represents the pixel number and $n(u)$ is the normal of the unit surface, which can be expressed based on the surface gradient:

$$n(u) = \frac{[-p(u), -q(u), 1]^T}{\sqrt{p(u)^2 + q(u)^2 + 1}} \quad (10)$$

where, $p(u) = \partial_x z(u)$ and $q(u) = \partial_y z(u)$.
The second linear constraint is obtained from the relation between unpolarized intensity and degree of polarization:

$$\frac{i_{un}(u)}{f(\rho(u),\eta)} = -p(u).s_x - q(u).s_y + s_z \quad (11)$$

in this equation, $s$ represents the light source.
Unknown height values can be obtained by a linear equation system with the least-squares method through gradient differences.

### 3.2 Height from polarization

Although the surface reconstructed by SfP method models the depth details in high resolution, it has a general error in the object's global curvature and three-dimensional geometry. Therefore, the polarization surface with high depth resolution is combined with the photogrammetric surface with high depth accuracy to solve this problem. In this way, the high-resolution/low-accuracy polarization surface is mounted on the low-resolution/high-accuracy photogrammetric surface, and their height calculations are performed pixel by pixel.

**3.2.1 Initial depth map:** This combination requires an initial depth map obtained from photogrammetric methods or laser scanners. The critical point in this section is that the image coordinate system between the three-dimensional photogrammetric and polarization model must be compatible. The multi-view stereo method (MVS) has been used to generate the initial photogrammetric depth model in this research.
In addition to the convergent images, the polarized images are also used in initial depth map generation to align the photogrammetric point cloud to the polarized image. The photogrammetric point cloud can be transferred from the object to the image space with interior and exterior orientation parameters.
According to the algorithm shown in Fig. 2 and Eq. (12), the photogrammetric point cloud is transferred from the object to the image space using the exterior orientation parameters of the polarized image. Next, the point cloud coordinates can be achieved in the polarized image coordinate system using the interior orientation parameters and applying lens radial and tangent distortion corrections (Eq. (13)).

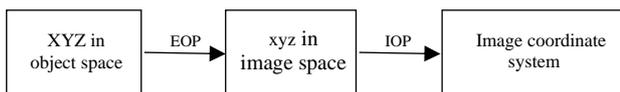

**Figure 2**. The transferring algorithm of photogrammetric point cloud to polarized image coordinate system

$$\begin{cases} x - x_0 + \Delta x = -f \frac{r_{11}(X-X_0)+r_{12}(Y-Y_0)+r_{13}(Z-Z_0)}{r_{31}(X-X_0)+r_{32}(Y-Y_0)+r_{33}(Z-Z_0)} \\ y - y_0 + \Delta y = -f \frac{r_{21}(X-X_0)+r_{22}(Y-Y_0)+r_{23}(Z-Z_0)}{r_{31}(X-X_0)+r_{32}(Y-Y_0)+r_{33}(Z-Z_0)} \end{cases} \quad (12)$$

In this equation, XYZ represent the object space coordinates of the measured points, and xyz are their image space coordinates. $(\Delta x, \Delta y)$ represent lens radial and tangent distortion, and the general model of distortion correction can be written in the following form:

$$\begin{cases} \Delta x = \bar{x}(k_1 r^2 + k_2 r^4 + k_3 r^6 + \cdots) + \cdots \\ \quad \ldots [p_1(r^2 + 2\bar{x}^2) + 2p_2 \bar{x}\bar{y}](1 + p_3 r^2 + \cdots) \\ \Delta y = \bar{y}(k_1 r^2 + k_2 r^4 + k_3 r^6 + \cdots) + \cdots \\ \quad \ldots [p_2(r^2 + 2\bar{y}^2) + 2p_1 \bar{x}\bar{y}](1 + p_3 r^2 + \cdots) \end{cases} \quad (13)$$

where $\bar{x} = x - x_0$, $\bar{y} = y - y_0$ and $r = \sqrt{\bar{x}^2 + \bar{y}^2}$.

**3.2.2 Post-processing:** After transferring the photogrammetric point cloud to the polarized image coordinate system, the value of x, y and z in each pixel are not unique, and there may be several unwanted invisible points in each pixel with different coordinates. The number of points in each pixel must be estimated first. Then each point distance from the center coordinate of the polarized image must be calculated, and the point with the shortest distance must be selected as visible one. By this processing, the points that are not seen in the image and located in hidden areas will be removed. Obviously, after transferring the photogrammetric point cloud to the polarized image coordinate system, several pixels have no coordinates, and no points are located in these pixels. On the other hand, because depth combination is done in each pixel separately, having three-dimensional coordinates in each pixel is necessary to perform calculations. In this step, the interpolation operation is performed to assign appropriate values to all pixels that do not have three-dimensional coordinates. After performing these two steps, the transferred photogrammetric point cloud can be used as an initial three-dimensional model to correct the three-dimensional geometry of the polarization surface.

**3.2.3 Combination method:** In order to combine these two models, firstly, the digital elevation model must be produced. Next, both models must be scaled in height to combine the two models correctly. In this way, both models are transferred to the range of 0-1, normalized, and then scaled. After performing these steps, the calculations can be performed correctly on the corresponding pixels of the photogrammetric and polarization surfaces.This combination method is performed to combine the photogrammetric and polarization surfaces based on the dimensions of the height grid. Depending on the selected grid dimensions, equations with unknown height differences will be created.

$$dz_{grid} = z_{MVS} - z_{polar} \quad (14)$$

As shown in Fig. 4, the photogrammetric surface contains three-dimensional geometry, and the polarized surface represents the details of the object. The size of both of these surfaces is the same as the polarized images' size ($m \times n$). Height differences are obtained using a 4-pixel grid. As the number of grid sizes decreases, the number of unknowns for registering heights increases, and the combination is performed at high frequencies. Next, a Gaussian filter is applied on height differences values for vertical smoothing on the grid. In the next step, the height differences are interpolated using the Bilinear method (Fig. 3) to obtain a specific height value for each pixel.

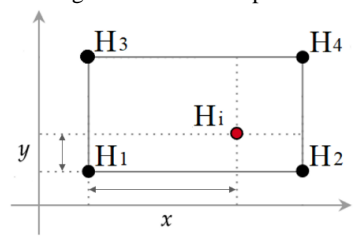

**Figure 3**. Grid cell design for height interpolation using bilinear method

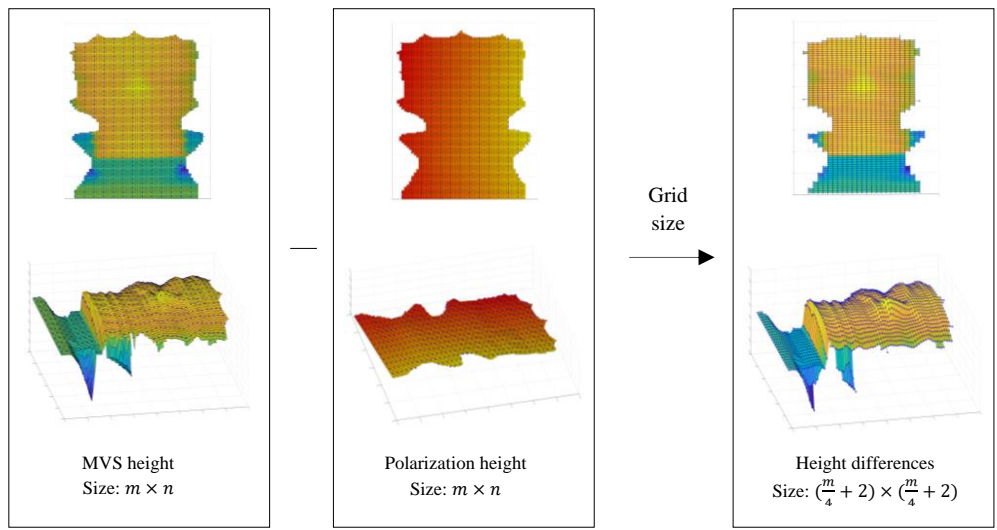

**Figure 4**. The process of obtaining height differences. The blue points on the surfaces indicate grid nodes

The bilinear function is usually written as:

$$H_i = a_{00} + a_{10}x + a_{01}y + a_{11}xy \qquad (15)$$

where:
$a_{00} = H_1$,
$a_{10} = H_2 - H_1$,
$a_{01} = H_3 - H_1$,
$a_{11} = H_1 - H_2 - H_3 + H_4$

Finally, the interpolated heights in each pixel are optimally added to the polarized heights.

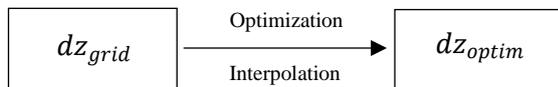

$$z_{combined} = z_{polar} + dz_{optim} \qquad (16)$$

After transferring the polarization height on the photogrammetric height using the bilinear method, the unknown height values will be obtained in the simultaneous optimization when the height differences between these two surfaces are minimized. In this case, projective transitions in each grid are applied at all polarization heights, and they will be most conforming to the photogrammetric surface.

After integration of photogrammetric and polarization surfaces, the combined surface is still in the image space. In order to perform quantitative assessments, it is necessary to transfer the combined surface to the object space. Cloud Compare software can be used for performing the transformation, and the object scale can be corrected by defining the exact length of the distances on the object.

## 4. EXPERIMENTAL RESULTS

### 4.1 Data collection and processing

In the implementation phase, the proposed method has been implemented on an image sequence captured by a Canon-EOS7D camera with a linear polarizing filter. The images have been captured at seven different angles (0, 30, 60, 90, 120 and 150 degrees) by filter rotation. The object and camera are fixed throughout the image capturing, and only the filter rotates.

The image registration must be implemented to reduce the errors created during image capturing due to camera slight movement. In this process, the first image taken at 0 angle has been considered the base image, and registration is done by the bicubic method under affine transformation model.

### 4.2 Polarization information and height reconstruction

After image registration, the preliminary information has been obtained from the polarized images according to the equations described in the previous section. The polarization information has been shown in Fig. 5, which includes unpolarized intensity, ($i_{un}$), degree of polarization, ($\rho$), phase angle, ($\varphi$), and zenith angle ($\theta$). The refractive index value in all experiments has been assumed to be 1.5. As shown in Fig. 5c, the phase angle obtained from this method is ambiguous, and its range varies between the values of $[0, \pi]$. It can also be seen that there are large phase deviations in some areas, despite the same neighborhoods. This issue confirms the ambiguity in the phase angle because both adjacent pixels should have approximately the same values. As mentioned earlier in the proposed method, phase ambiguity is removed simultaneously with the calculation of height values.

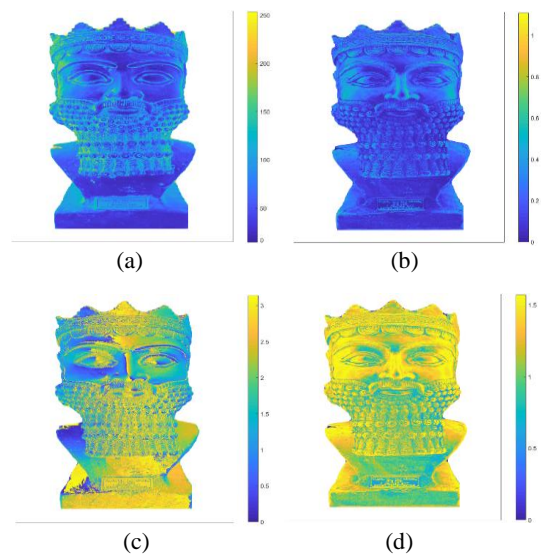

**Figure 5**. (a) Unpolarized intensity light; (b) Degree of polarization; (c) Phase angle and (d) Zenith angle.

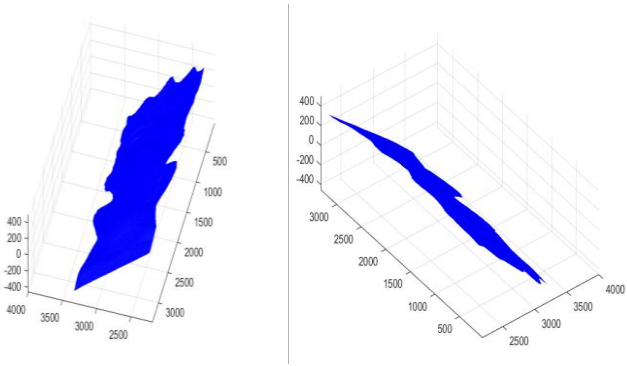

**Figure 6**. Height obtained from the polarization method before curvature correction

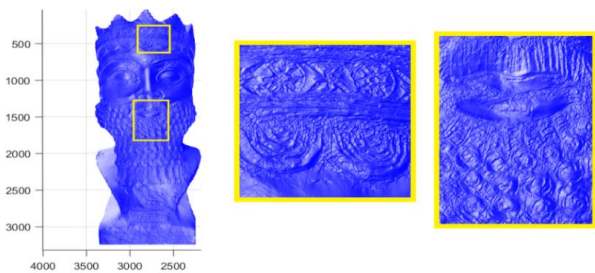

**Figure 7**. Details from the polarization method before curvature correction.

Although the reconstructed surface is almost flat, it has been able to reconstruct the details in these objects well (see Fig. 6 and 7). In order to evaluate the SfP method performance on objects with different materials, a glassy object and a shiny metal coin have also been examined. As shown in Fig. 8, this method can reconstruct all the details of glassy and shiny objects.

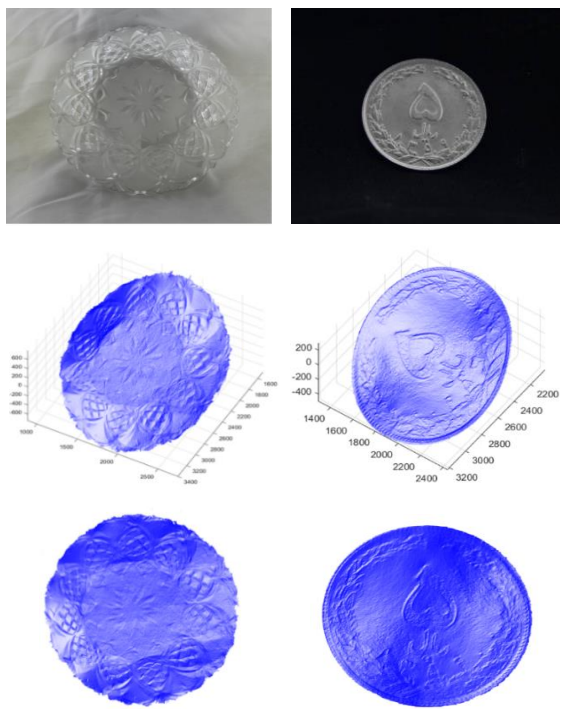

**Figure 8**. Polarized image and depth obtained from the polarization method in glassy object and a coin before curvature correction

### 4.3 Initial depth map and combination

After capturing convergent images and producing a three-dimensional model, the photogrammetric point cloud is transferred to the polarized image coordinate system using interior and exterior orientation parameters (Eq. (12 and 13)). The green dots in Fig. 9 represent the transferred points cloud to the polarized image system. After post-processing and producing a digital elevation model, both models are ready for combination. To have a high accuracy high resolution surface reconstruction, this paper has done the combination process with the smallest grid size, 4, after height scaling. As shown in Fig. 10, the three-dimensional geometry in the combined surface corresponds to the photogrammetric surface, and also all the details in the combined surface are well observed in different views.

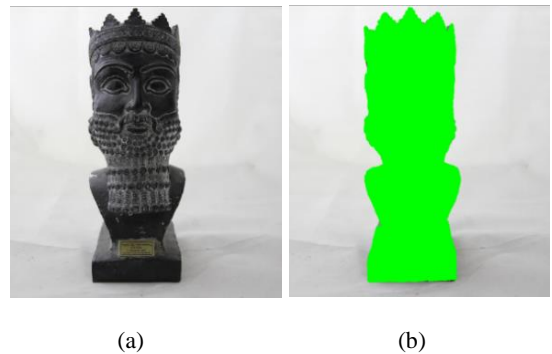

(a)          (b)

**Figure 9**. (a) polarized image and (b) the location of the photogrammetric points cloud on the polarized image

### 4.4 Assessments

**4.4.1 Visual assessment:** Three profiles on the combined surface of the cultural heritage object have been considered for visual assessment (Fig. 11). These profiles have been compared between the proposed and conventional photogrammetric methods at the corresponding surfaces. The distance of the profiles shown in Fig. 12a and Fig. 12b is about one centimeter. They show that the profiles obtained from the reconstructed surface based on the shape from polarization method have more details than the MVS method and all curvatures, tracks and valleys below the millimeter are well modelled.

For visual assessment of the smooth and non-detail surface reconstruction performance in the SfP method compared to the MVS photogrammetric method, profile number 3 has been considered (Fig. 13). As shown in Figure 13, the reconstructed profile in SfP method is approximately similar to MVS method. This means for smooth surfaces, based on the sampling theory, it does not need to combine a high-resolution surface reconstruction method which here is SfP with photogrammetry. As profile 3 gets from a smooth area, its RMSE shows the basic noise of proposed method. In rough surfaces with high details, such as profiles 1 and 2, the RMSE, in fact, shows the result noise of photogrammetry and polarization in reconstruction of high frequency details. By error propagation rule, the second row shows the error of photogrammetry method. The row 3 shows relative error of proposed method to photogrammetry method which says the proposed method has 2.7 times higher reconstruction accuracy than conventional photogrammetry method.

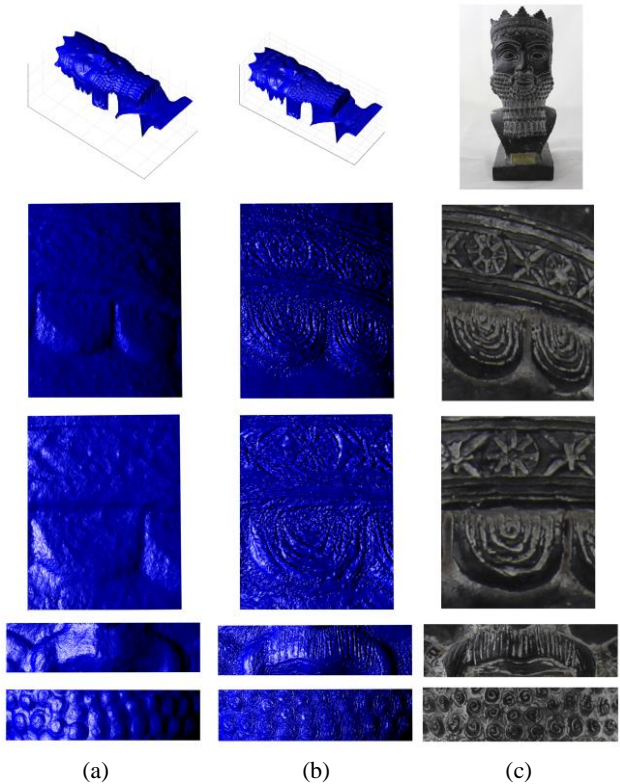

(a) (b) (c)

**Figure 10**. (a) photogrammetric surface from MVS method; (b) combined surface and (c) image of the patches

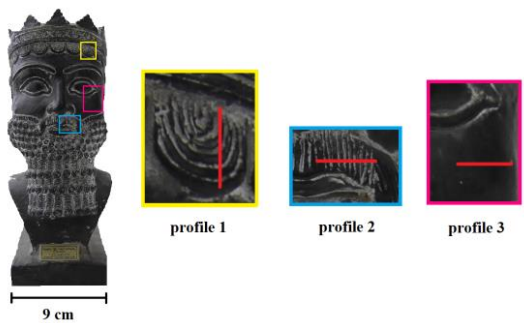

**Figure 11**. Geometric location of three profiles on the three-dimensional model

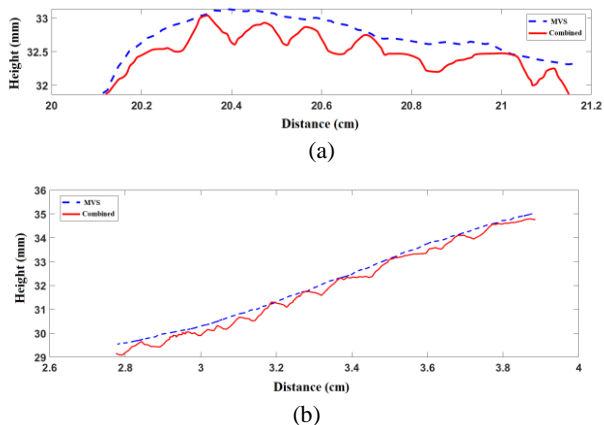

**Figure 12**. (a) profile 1 and (b) profile 2

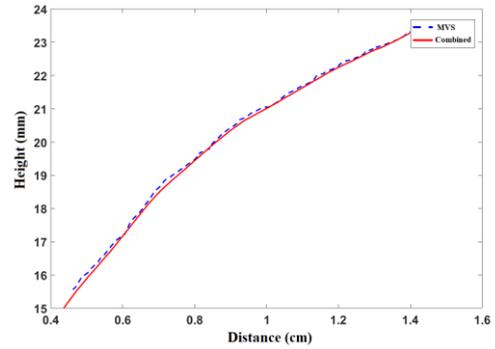

(c)
**Figure 13**. profile 3

|  | profile 1 | profile 2 | profile 3 |
|---|---|---|---|
| **RMSE (mm)** | 0.2676 | 0.2618 | 0.0924 |
| **Photogrammetry RMSE (mm)** | 0.2511 | 0.2450 | - |
| **Relative Error** | 1:2.72 | 1:2.65 | - |

**Table 1**. The values of Root-Mean-Square Error between photogrammetry and polarization methods

**4.4.2 Quantitative assessment:** High accurate smooth and flat surfaces in different objects have been considered as a geometric reference to investigate the relative error and noise of the proposed method by fitting a flat plane to these surfaces. A flat area has been considered on the golden label in the Dariush statue to estimate the error, and other flat areas on the coin and the glassy object, shown with a red square in Fig. 14. In the next step, a flat plane has been fitted on these areas to calculate the root-mean-square error of the combined surface relative to the fitted plane. The values of root-mean-square error relative to GSD in three objects have been shown in Table 2.

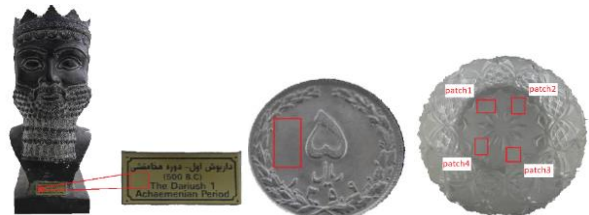

**Figure 14**. Selected patches to calculate the relative error value and noise in the proposed method

|  | Dariush statue | Coin | Glassy object |
|---|---|---|---|
| **RMSE (mm)** | 0.0486 | 0.0028 | 0.0059 |
| **GSD (mm)** | 0.17 | 0.019 | 0.07 |
| **GSD/ RMSE** | 0.285 | 0.147 | 0.084 |

**Table 2**. The values of root-mean-square error and GSD in the Dariush statue, coin and glassy object

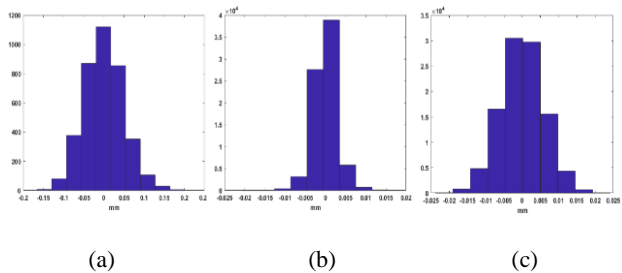

(a)　　　　　　(b)　　　　　　(c)

**Figure 15**. Histogram of the height differences values between the fitted plane and the combined surface in (a) Dariush statue; (b) coin and (c) glassy object.

Our experiments show that this method can reconstruct depth details of different surfaces with very high resolution and in the range of 0.1 to 0.3 pixels for the objects smaller than 50 centimeters which is roughly 10 times better than conventional photogrammetric method.

## 5. CONCLUSION

This paper presents a combined solution to increase depth resolution, with significant advantages over conventional photogrammetric methods. This proposed method uses polarization information and Fresnel theory to reconstruct surface details and combine them with the multi-view stereo method. The surface details are first obtained using the polarization information in this solution. Then, the curvature correction is done using an initial depth map to modify the model in terms of three-dimensional geometry. Depth maps can be significantly enhanced by using the information obtained from the polarization of light. Visual assessments of a cultural heritage object's reconstructed surface in profiles of Fig. 12 demonstrate this method's ability to reconstruct surface details below millimeters and its capacity to improve photogrammetric depth map resolution using the polarization method's details. In quantitative assessment, the method's relative errors in different objects have been estimated by fitting a flat plate. The RMSE values in the statue of Dariush, coin and glassy object are 0.0486, 0.0028 and 0.0059 mm, respectively. According to the value of GSD in these objects, the errors rate per pixel are equal to 0.285, 0.147 and 0.084, respectively.

The results of visual and quantitative assessments showed that the proposed method could reconstruct details more petite than a millimeter in cultural heritage objects with dimensions less than 50 centimeters. It should be noted that the main advantage of this method is the ability to measure and estimate the depth of glassy and shiny metal objects as well as featureless objects that photogrammetric methods have difficulty in reconstructing.

Finally, it is emphasized again that the most significant benefit of this method is that it is a passive method and does not require controlled lighting and can be considered a complementary method in close-range photogrammetry and even UAV photogrammetry.

## REFERENCES

Atkinson, G.A. and Ernst, J.D., 2018. High-sensitivity analysis of polarization by surface reflection. Machine Vision and Applications, 29(7), pp.1171-1189.

Atkinson, G.A. and Hancock, E.R., 2005, October. Multi-View Surface Reconstruction Using Polarization. In ICCV (Vol. 2, p. 3).

Atkinson, G.A. and Hancock, E.R., 2006. Recovery of surface orientation from diffuse polarization. IEEE transactions on image processing, 15(6), pp.1653-1664.

Atkinson, G.A. and Hancock, E.R., 2007a, August. Surface reconstruction using polarization and photometric stereo. In International conference on computer analysis of images and patterns (pp. 466-473). Springer, Berlin, Heidelberg.

Atkinson, G.A. and Hancock, E.R., 2007b. Shape estimation using polarization and shading from two views. IEEE transactions on pattern analysis and machine intelligence, 29(11), pp.2001-2017.

Atkinson, G.A., 2017. Polarisation photometric stereo. Computer Vision and Image Understanding, 160, pp.158-167.

Ba, Y., Gilbert, A., Wang, F., Yang, J., Chen, R., Wang, Y., Yan, L., Shi, B. and Kadambi, A., 2020, August. Deep shape from polarization. In European Conference on Computer Vision (pp. 554-571). Springer, Cham.

Collett, E., 2005, September. Field guide to polarization. Bellingham, WA: Spie.

Cui, Z., Gu, J., Shi, B., Tan, P. and Kautz, J., 2017. Polarimetric multi-view stereo. In Proceedings of the IEEE conference on computer vision and pattern recognition (pp. 1558-1567).

Deschaintre, V., Lin, Y. and Ghosh, A., 2021. Deep polarization imaging for 3D shape and SVBRDF acquisition. In Proceedings of the IEEE/CVF Conference on Computer Vision and Pattern Recognition (pp. 15567-15576).

Drbohlav, O. and Sara, R., 2001, July. Unambiguous determination of shape from photometric stereo with unknown light sources. In Proceedings Eighth IEEE International Conference on Computer Vision. ICCV 2001 (Vol. 1, pp. 581-586). IEEE.

Esteban, C.H., Vogiatzis, G. and Cipolla, R., 2008. Multiview photometric stereo. IEEE Transactions on Pattern Analysis and Machine Intelligence, 30(3), pp.548-554.

Haque, M., Chatterjee, A. and Madhav Govindu, V., 2014. High quality photometric reconstruction using a depth camera. in proceedings of the IEEE conference on computer vision and pattern recognition (pp. 2275-2282).

Hecht, E., 2002. Optics, 4th editio ed. Addison-Wesley, San Francisco, 2, p.3.

Huynh, C.P., Robles-Kelly, A. and Hancock, E., 2010, June. Shape and refractive index recovery from single-view polarisation images. In 2010 IEEE Computer Society Conference on Computer Vision and Pattern Recognition (pp. 1229-1236). IEEE.

Kadambi, A., Taamazyan, V., Shi, B. and Raskar, R., 2017. Depth sensing using geometrically constrained polarization normals. International Journal of Computer Vision, 125(1), pp.34-51.

Kalra, A., Taamazyan, V., Rao, S.K., Venkataraman, K., Raskar, R. and Kadambi, A., 2020. Deep polarization cues for transparent object segmentation. In Proceedings of the IEEE/CVF Conference on Computer Vision and Pattern Recognition (pp. 8602-8611).

Langguth, F., Sunkavalli, K., Hadap, S. and Goesele, M., 2016, October. Shading-aware multi-view stereo. European Conference on Computer Vision (pp. 469-485). Springer, Cham.

Lei, C., Qi, C., Xie, J., Fan, N., Koltun, V. and Chen, Q., 2021. Shape from Polarization for Complex Scenes in the Wild. arXiv preprint arXiv:2112.11377.

Mahmoud, A.H., El-Melegy, M.T. and Farag, A.A., 2012, September. Direct method for shape recovery from polarization and shading. In 2012 19th IEEE International Conference on Image Processing (pp. 1769-1772). IEEE.

Mecca, R., Logothetis, F. and Cipolla, R., 2017. A differential approach to shape from polarization.

Mitra, N.J. and Nguyen, A., 2003, June. Estimating surface normals in noisy point cloud data. In Proceedings of the nineteenth annual symposium on Computational geometry (pp. 322-328).

Miyazaki, D., Saito, M., Sato, Y. and Ikeuchi, K., 2002. Determining surface orientations of transparent objects based on polarization degrees in visible and infrared wavelengths. JOSA A, 19(4), pp.687-694.

Miyazaki, D., Shigetomi, T., Baba, M., Furukawa, R., Hiura, S. and Asada, N., 2016. Surface normal estimation of black specular objects from multiview polarization images. Optical Engineering, 56(4), p.041303.

Miyazaki, D., Shigetomi, T., Baba, M., Furukawa, R., Hiura, S. and Asada, N., 2012, October. Polarization-based surface normal estimation of black specular objects from multiple viewpoints. In 2012 Second International Conference on 3D Imaging, Modeling, Processing, Visualization & Transmission (pp. 104-111). IEEE.

Miyazaki, D., Tan, R.T., Hara, K. and Ikeuchi, K., 2003, October. Polarization-based inverse rendering from a single view. In Computer Vision, IEEE International Conference on (Vol. 3, pp. 982-982). IEEE Computer Society.

Moons, T., Van Gool, L. and Vergauwen, M., 2010. 3D reconstruction from multiple images part 1: Principles. Foundations and Trends® in Computer Graphics and Vision, 4(4), pp.287-404.

Morel, O., Meriaudeau, F., Stolz, C. and Gorria, P., 2005, February. Polarization imaging applied to 3D reconstruction of specular metallic surfaces. In Machine Vision Applications in Industrial Inspection XIII (Vol. 5679, pp. 178-186). International Society for Optics and Photonics.

Nehab, D., Rusinkiewicz, S., Davis, J. and Ramamoorthi, R., 2005. Efficiently combining positions and normals for precise 3D geometry. ACM transactions on graphics (TOG), 24(3), pp.536-543.

Ngo Thanh, T., Nagahara, H. and Taniguchi, R.I., 2015. Shape and light directions from shading and polarization. In Proceedings of the IEEE conference on computer vision and pattern recognition (pp. 2310-2318).

Oxholm, G. and Nishino, K., 2014. Multiview shape and reflectance from natural illumination. Proceedings of the IEEE Conference on Computer Vision and Pattern Recognition (pp. 2155-2162).

Park, J., Sinha, S.N., Matsushita, Y., Tai, Y.W. and Kweon, I.S., 2013. Multiview photometric stereo using planar mesh parameterization. Proceedings of the IEEE International Conference on Computer Vision (pp. 1161-1168).

Polarization_camera, 2020: https://en.ids-imaging.com/techtipp-details/techtip-on-camera-polarization.html (Accessed 14st April 2022).

Smith, W.A., Ramamoorthi, R. and Tozza, S., 2016, October. Linear depth estimation from an uncalibrated, monocular polarisation image. In European Conference on Computer Vision (pp. 109-125). Springer, Cham.

Smith, W.A., Ramamoorthi, R. and Tozza, S., 2018. Height-from-polarisation with unknown lighting or albedo. IEEE transactions on pattern analysis and machine intelligence, 41(12), pp.2875-2888.

Tozza, S., Smith, W.A., Zhu, D., Ramamoorthi, R. and Hancock, E.R., 2017. Linear differential constraints for photo-polarimetric height estimation. In Proceedings of the IEEE international conference on computer vision (pp. 2279-2287).

Wolff, L.B. and Boult, T.E., 1993. Constraining object features using a polarization reflectance model. Phys. Based Vis. Princ. Pract. Radiom, 1, p.167.

Wolff, L.B., 1997. Polarization vision: a new sensory approach to image understanding. Image and Vision computing, 15(2), pp.81-93.

Wu, C., Wilburn, B., Matsushita, Y. and Theobalt, C., 2011, June. High-quality shape from multi-view stereo and shading under general illumination. In CVPR 2011 (pp. 969-976). IEEE.

Yang, L., Tan, F., Li, A., Cui, Z., Furukawa, Y. and Tan, P., 2018. Polarimetric dense monocular SLAM. In Proceedings of the IEEE conference on computer vision and pattern recognition (pp. 3857-3866).

Yu, Y., Zhu, D. and Smith, W.A., 2017. Shape-from-polarisation: a nonlinear least squares approach. In Proceedings of the IEEE International Conference on Computer Vision Workshops (pp. 2969-2976).

Zhou, Z., Wu, Z. and Tan, P., 2013. multi-view photometric stereo with spatially varying isotropic materials. proceedings of the IEEE conference on computer vision and pattern recognition (pp. 1482-1489).

Zhu, D. and Smith, W.A., 2019. Depth from a polarisation+ RGB stereo pair. In Proceedings of the IEEE/CVF Conference on Computer Vision and Pattern Recognition (pp. 7586-7595).